\pdfoutput=1
\documentclass[11pt]{article}

\usepackage[preprint]{acl}
\usepackage{booktabs}
\usepackage{times}
\usepackage{latexsym}
\usepackage{booktabs}
\usepackage[T1]{fontenc}
\usepackage[utf8]{inputenc}

\usepackage{microtype}

\usepackage{inconsolata}

\usepackage{graphicx}

\newcommand{\numAnnotated}{400}

\newcommand{\numArticlesAnnotated}{836}

\newcommand{\numsamples}{7,774}
\newcommand{\nummulti}{3,869}
\newcommand{\numhighlights}{597}
\newcommand{\numparagraph}{4,277}
\newcommand{\numsentence}{2,900}

\newcommand{\fpt}[0]{front-page teasers}

\newcommand{\FPT}[0]{Front-Page Teasers}
\newcommand{\name}[0]{\textsc{HebTeaseSum}}

\title{Leveraging Digitized Newspapers to Collect Summarization Data in Low-Resource Languages}

\author{Noam Dahan \quad Omer Kidron \quad Gabriel Stanovsky \\[1.5mm]
         The Hebrew University of Jerusalem \\[1.5mm] 
 \href{mailto:noam.dahan1@mail.huji.ac.il}{\texttt{noam.dahan1@mail.huji.ac.il}}}

\usepackage[normalem]{ulem}

\newcommand{\com}[1]{}

\newcommand{\resolved}[1]{}

\begin{document}
\maketitle
\begin{abstract}
High quality summarization data remains scarce in under-represented languages. However, historical newspapers, made available through recent digitization efforts, offer an abundant source of untapped, naturally annotated data. In this work, we present a novel method for collecting naturally occurring summaries via \emph{\FPT{}}, where editors  summarize full length articles. We show that this phenomenon is common across seven diverse languages and supports multi-document summarization. To scale data collection, we develop an automatic process, suited to varying linguistic resource levels. Finally, we apply this process to a Hebrew newspaper title, producing \name{}, the first dedicated multi-document summarization dataset in Hebrew\footnote{\url{https://github.com/edahanoam/HebTeaseSum}}.

\end{abstract}

\section{Introduction}

%While recent studies suggest that LLMs may outperform existing summarization benchmarks in English \cite{zhang2024benchmarking} so much so that \citet{pu2023summarizationalmostdead}, declared this task to be "almost dead", non-English languages suffer from lack of accessible, high-quality datasets \cite{dahan-stanovsky-2025-state}. This limits the field's ability to evaluate model performance across many languages and hinders progress in research. 

Recent studies suggest that the task of summarization in English may be already solved, or even ``(almost) dead''~\cite{pu2023summarizationalmostdead,zhang2024benchmarking}. However, this is not the case in the vast majority of world languages, which suffer from lack of accessible, high-quality summarization datasets~\cite{dahan-stanovsky-2025-state}.
% , which limits the ability to evaluate and develop models for the vast majority of languages.
Collecting high-quality summarization data in a new language is hard: human annotation in summarization is particularly challenging~\cite{varab2021massivesumm}, while automatic data collection methods depend on the availability of web content, which is frequently limited in low-resource settings~\cite{joshi2020state}.
%\gabis{I changed this paragraph a bit, previous version is in comments. Unclear what it means to outperform existing benchmarks}

\begin{figure}[tb!]
  \includegraphics[width=0.85\columnwidth]{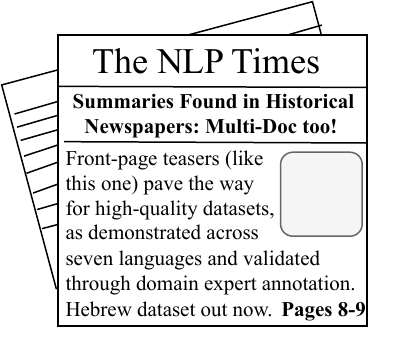}
  \centering
  \caption{Newspaper's \fpt{} are a natural source for high-quality, expert-written summaries of full articles appearing inside the paper. We show that they are common across languages, and lend themselves to straightforward data collection.}
  \label{fig:figur1}
\end{figure}

%\gabist{In this work, we present a \gabisrep{novel}{straightforward and scalable} method for collecting naturally annotated summarization data based on \gabisrep{printed}{digitized print} newspapers \gabist{rather than online data.} \gabisrep{It requires}{requiring} minimal human annotation and \gabist{is} suitable for various \gabisrep{resource levels}{low-resource languages}.} \gabis{we present more than a method...}
\begin{figure*}[tb!]
    \includegraphics[width=\textwidth]{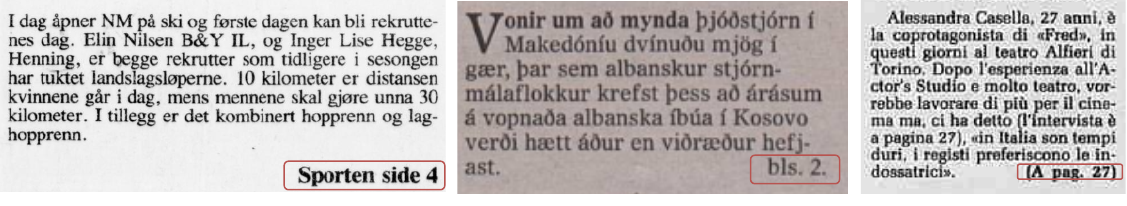}
    \caption{\textbf{\FPT{} are common in a variety of languages}. Highlighted text shows the reference to relevant pages where the corresponding articles can be found, and  serves as a useful signal for identifying summaries. From Left-to-Right: teasers from Rana Blad (Norway), Fréttablaðið (Iceland) and Stampa Sera (Italy). Links to newspapers and translations are in Appendix \ref{sec:appendixn}.}
    %\caption{\textbf{\FPT{} are high-quality summaries, common in a variety of languages}. Highlighted text shows the reference to relevant pages where the corresponding articles can be found, and is a good indication for identifying summaries. Images showcase the languages for which we collect data. From top-to-bottom, left-to-right: teasers from Rana Blad (Norway), Kathimerini (Greece), Stampa Sera (Italy), Eesti Päevaleht (Estonia), Fréttablaðið (Iceland), Dziennik Polski (Poland) and Maariv (Israel). Links to newspapers are in Appendix \ref{sec:appendixn}.}
      \label{fig:example_multiling}
\end{figure*}

In this work, we observe that traditional print newspapers contain  organic, high-quality summarization data. In particular, we leverage \emph{\fpt{}}, short blurbs written by professional editors describing one or more articles appearing inside the issue~\cite{utt1989they}, as illustrated in Figure~\ref{fig:figur1}. We will show that \fpt{} constitute high-quality, expert-written summaries, which are abundant in a  range of languages.

% Our work is driven by the observation that traditional print newspapers contain  often feature \emph{\fpt{}}, these are \gabisrep{summaries}{blurbs} written by professional editors describing one or more articles appearing inside the issue~\cite{utt1989they}, as illustrated in Figure~\ref{fig:figur1}. 

% In Section~\ref{sec:Background}, we describe the growing availability of historical newspapers, thanks to global digitization efforts and advances in OCR technology, making \fpt{} a plentiful resource for high-quality summarization signal across many languages. 

%In Section~\ref{sec:method_new} we describe how to gather data for the summarization task through \fpt{}, using a simple two step process. We show that teasers are common by manually collecting small scale data in seven diverse languages. We further show they are highly abstractive and support multi-document summarization, which occur when a single front-page teaser refers to multiple news stories. \noamd{change this part to include human annotation}

% In Section~\ref{sec:method_new}, 
We begin by demonstrating that \fpt{} can be collected in a simple, two-step process, and show that they are widely available by collecting data in seven diverse language. Furthermore, we validate the suitability of teasers as summaries through expert annotation, which found them to be high quality and consistent with the source document. Interestingly, we find that \fpt{} are also a rich source for \emph{multi-document summarization}, when a single \fpt{} summarizes several news stories.

% we describe how to gather data for the summarization task through \fpt{}, using a simple two step process. 
% First, we show that teasers are common by manually collecting small scale data in seven diverse languages, and often support multi-document summarization, which occur when a single front-page teaser refers to multiple news stories. Second, we validate the suitability of teasers as summaries through expert annotation, which found them to be high quality and consistent with the source document.%, with length indicating the amount of important information included. 

% In Section~\ref{sec:Experiments}, 
Following, we use \fpt{} to evaluate state-of-the-art LLMs on summarization across diverse languages. We find that LLMs struggle to fully cover all of the information in the gold summary. Moreover, performance disparities between models are significantly larger in lower-resource languages, underscoring the need to curate high-quality datasets in these settings.

Finally, we show that \fpt{} are also a source for large scale evaluation or fine tuning. 
Using a simple heuristic over a large corpus of Hebrew print newspapers, we create \name{}, the first dedicated summarization dataset for Hebrew which supports multi-document summarization, totaling \numsamples{} samples.
% We develop a simple automatic method which automatically matches between \fpt{} and corresponding articles.  
% We make sure our approach is suited to a wide range of languages by conducting experiments with different methods to make this alignment ranging from LLMs to corropus statistics methods, such as TF-IDF. 

To conclude, we make the following contributions:
\begin{itemize}
    \item We introduce a novel approach for building summarization datasets, using untapped naturally annotated data of historical newspapers.
    %\item We show that our method enables data collection in many languages, and provide a first summarization evaluation of state-of-the-art models on a wide range of languages.
    \item We show that our approach is applicable to many languages and supports model evaluation across varying linguistic resource levels.
    \item We release a multi-document summarization dataset in Hebrew, created using our proposed approach.
\end{itemize}

%\noamd{could be nice to talk about how people are moving to synthetic data thinking there is no more availabe data - but this is an amazing reousrce that it not explored just yet}

%this from summarization is almost dead:
%In order to thoroughly assess the
%summarization capabilities of LLMs, it becomes
%imperative to incorporate other diverse genres of
%data, as well as other languages, especially those
%that are low-resource in nature. Additionally, there
%is a need to include longer documents, such as
%books, within the datasets to facilitate comprehensive evaluation.

\section{Background}
\label{sec:Background}
\begin{figure*}[tb!]
  \includegraphics[width=\textwidth]{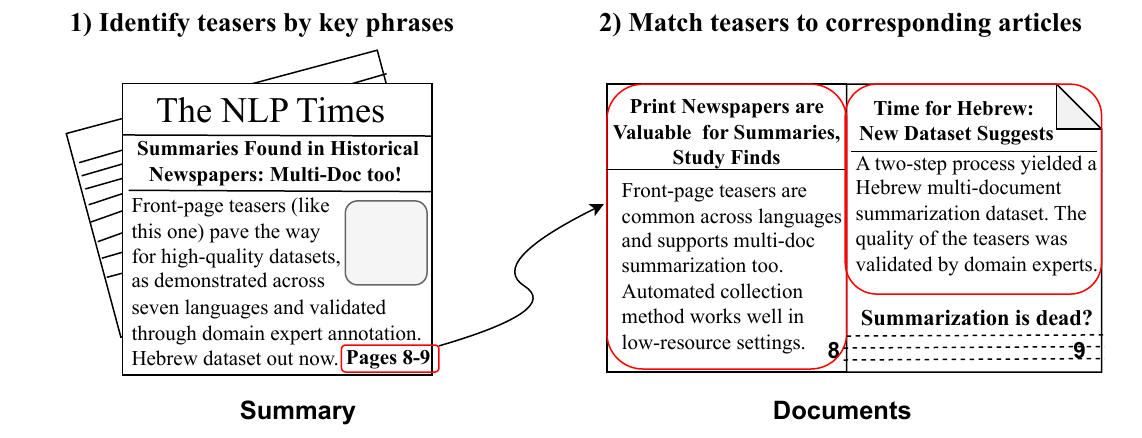}
  \caption{\textbf{Our approach for extracting a summarization dataset from printed newspapers.} We first find teasers and corresponding page numbers on the front page based on newspaper-specific keywords (e.g., ``Full articles on Pages 8-9''). Then, we turn to these pages and find the articles most relevant to the content of the teaser, thus resulting in (teaser summary, relevant articles) pairs.}
  \label{fig:fullMethod}
\end{figure*}

In recent years, extensive efforts have been made to digitize historical newspapers, making them increasingly accessible for NLP research. In this section, we describe this growing resource and clarify the journalism terminology used in this work.

\paragraph{Extensive Digitization Efforts.} National libraries world wide preserve  collections of historical newspapers in many languages~\cite{beals2020atlas}. The library of the US congress alone curates texts in 36 languages including native languages considered to be endangered~\cite{zhang2022can}. Similar multilingual newspaper archives can be found across Europe as well as in national libraries of at least 163 countries \cite{haneefa2019contents}. 

These newspapers are becoming increasingly accessible to researchers due to vast digitization and open access efforts%such as \cite{beals2020atlas}
. 
In 2015, it was estimated that there are more than 45  thousand digitized newspaper titles, while the actual number may be significantly higher \cite{kettunen2020ground}. 
This is especially important for low resource languages, as many of them have minimal web presence~\cite{joshi2020state}.

Moreover, recent advances in OCR models such as Mistral OCR and improved post-correction methods \cite{thomas2024leveraging} give hope for overcoming past challenges in using this valuable data, especially the problems caused by errors in text recognition and segmentation \cite{dell2023american}.

\paragraph{Terminology.} We define key terms we use in this work. \emph{Newspaper titles} refer to publications such as ``The New York Times'' or ``The Daily Mail'', while an \emph{issue} refers to a specific edition of a newspaper title, for example, the issue of The New York Times that came out on January 1st, 1990. We  refer to the short text appearing on the front page and pointing to an article on a different page within the issue as a \emph{front-page teaser}. These are also known as `lead-in' or 'synopsis' \cite{utt1989they}. Front page teasers are different from \emph{leads} of articles sometimes appearing on the front page, that contain the \emph{beginning} of an article and point the reader to its continuation elsewhere in the issue.

%These rich cultural text has been recognize as a valuable source in social sciences \cite{soffer2020computational,}, yet they remain underexplored for NLP downstream tasks. 

% \section{Terminology}
% \label{sec:Terminology}
% \input{sections/Terminology}                                               

% \section{Targeted Collection For Evaluation}
% \label{sec:Data}
% \input{latex/sections/Data}                       
\section{Front-Page Teasers are Common, High-Quality Summaries}
\label{sec:method_new}
In this section we describe our approach to extract a summarization dataset given a collection of printed newspapers. We conduct  human evaluation performed by domain experts, to demonstrate \fpt{} high quality. Then, we show the organic front-page summaries heuristic holds across several languages with various levels of resources, and naturally supports multi-document summarization.

% In this section we describe our approach to extract a summarization dataset given a collection of printed newspapers. Then, through collection of multilingual data in a small scale, we show the organic front-page summaries heuristic holds across several languages with various levels of resources, and naturally supports multi-document summarization.  

%In this section we describe our approach to extract a summarization dataset given a collection of printed newspapers, as illustrated in Figure \ref{fig:fullMethod}. Then, through collection of multilingual data in a small scale, we show the organic front-page summaries heuristic holds across several languages with various levels of resources, and naturally supports multi-document summarization.  

\subsection{Constructing Summarization Data from Teasers}

Newspapers often provide readers with front-page teasers, summarizing the main points of one or more articles in the issue, see Figure \ref{fig:example_multiling} for examples. These teasers have been popular for at least the last 50 years~\cite{utt1989they}. Driving our work is the observation that front-page teasers are high-quality, multi-document summaries which can provide valuable signal for low resource languages. 

To collect a dataset reflecting that observation, the process consists of two steps, as shown in Figure~\ref{fig:fullMethod}: First, identifying the teasers, and second, matching them to the articles to which they point. 

Identifying front-page teasers is a simple process that requires determining the key terms each newspaper title uses to indicate that a full story is available in the issue. For example, teasers often end with a phrase like \emph{``page 5''}, indicating where the full article appears.

The second step, after identifying the \fpt{}, is to turn to the relevant pages to locate the full articles related to it. This step is more challenging than the first, for several reasons: First, the teaser and the article often have different headlines. Second, the same page may contain articles on similar topics. Finally, we find that \fpt{} rarely mentioned how many articles the summary supports. Nevertheless, a careful reading allows readers to identify the relevant articles.

Following these two steps, produces a collection of articles and the teasers summarizing them in an abstractive and organic manner. %In the following section, we will show that this phenomenon is indeed common across languages, and is widely accessible through national digitization efforts.

\subsection{Human Evaluation}
To assess \fpt{} suitability as summaries, we conduct human evaluation by two domain experts, each with over five years of experience in journalism. Their analysis, detailed in this section, showed that teasers are generally of high quality, with length serving as a useful indicator. Results are presented in Table~\ref{tab:human}. 

We randomly select \numAnnotated{} samples from the Hebrew newspaper used in our dataset along with \numArticlesAnnotated{} corresponding articles. For a meaningful evaluation, we divide the teasers into four  groups based on length and ensured an equal number of samples (100) from each category. A translated example for each of the categories is provided in Appendix~\ref{sec:appendixe}.

We follow~\citet{fabbri2021summeval} to score each teaser according to four metrics: (1) \emph{Coherence}, measuring how well the sentences in the summary fit together and whether they sound natural; (2) \emph{Consistency}, measuring whether the facts in the summary are consistent with the facts in the original article; (3) \emph{Fluency}, measuring quality of each sentence individually and (4)  \emph{Relevance}, measuring how much of the important information appears in the summary.\footnote{Annotator guideline can be found in Appendix~\ref{sec:appendixh}} 
%Our goal was to determine whether teasers exhibit shortcomings previously identified in automatically gathered, web-sourced summaries from the news domain, and whether low-quality teasers -- if exists -- can easily be filtered out. Specifically, we aim to ensure that teasers are informative rather than merely designed to lure readers into the article, a problem noted by \citet{lee2022headlines}, and that they do not include information unsupported by the article itself, as reported by \citet{tejaswin2021well}. These concerns are addressed through the relevance and consistency annotations, respectively.
Results are presented in Table~\ref{tab:human}. To measure inter-annotator agreement, the annotators overlapped on 100 teaser annotations (25 per category), achieving a Krippendorff’s alpha~\cite{krippendorff2018content} of 0.74, an agreement level often achieved in similar tasks~\cite{antoine2014weighted}. %\gabis{$\leftarrow$ rephrase this sentence to start with "To measure inter-annotator agreement, the annotators overlapped in annotation over ...} \gabis{Also, do we have something to say about this 0.64$\alpha$? Is this considered good? There are usually agreed upon spans for different agreement scores.}

All length categories achieved high scores in fluency and coherence, as they were written by professional editors to be featured on the front-page. The most common cause of point deduction in these categories was the use of ``bullet style'' sentences, which, while grammatically correct, were perceived as less natural by our annotators.

Moreover, the teasers were found to be consistent with the source document. In all length spans we found that \fpt{} contain information that is present in the source articles. This serves as a valuable indicator of the quality of teasers as summaries, especially given that web sourced summaries from the news domain were previously found to contain information that does not appear in the articles they are based on \cite{tejaswin2021well}. The difference may lie in the fact that, while web-scraping heuristics extract parts of the article (such as the first sentence or the title) to be used as summaries, teasers are intended to be read independently from the news article as they appear on a different page. 

The Relevance score exhibited the most score variance, with length clearly influencing how much of the important information was included in the teaser. This is most prominent when the teaser is under 25 words, which typically consisted of a single sentence.%This aligns with previous work on headlines, which sometimes aim to lure readers into the article rather than provide useful information~\cite{lee2022headlines}. 
 %\gabist{This shortcoming is addressed by removing shorter teasers, except when used for the extreme Summarization task \cite{narayan2018don}, which requires short summaries.}

\begin{table}[tb!]
\small
\begin{tabular}{@{}l|llll@{}}
\toprule
Len.            & Coherence & Consistency & Fluency & Relevance \\ \midrule
0-25     & 4.31      & 4.75        & 4.36    & 3.29      \\
25-50     & 4.47     & 4.73       & 4.74    & 3.90     \\
50-100    & 4.63     & 4.68        & 4.91    & 4.20      \\
>100 & 4.88       & 4.32        & 4.98    & 4.30      \\ \bottomrule
\end{tabular}
\caption{\textbf{Average teaser scores provided by domain expert annotators.} ``Len.''' denotes teaser length (in words). Relevance increases with teaser length while other scores are generally high.}
\label{tab:human}
\end{table}

\subsection{Front-Page Teasers are Common in a Wide Range of Languages}

\begin{table*}[tb!]
% \small

\centering

%\resizebox{\textwidth}{!}{

\begin{tabular}{@{}lllllcccc@{}l}
\toprule
\textbf{Language} &\textbf{Res.}$^*$ & \textbf{Title}&\textbf{Comp.}$^{**}$ & \multicolumn{4}{c}{\textbf{Novel n-grams}} & \textbf{\%Multi-doc}$^{***}$\\
\cmidrule(lr){5-8}
         &  & &     & 1 & 2 & 3 & 4 \\
\midrule
Norwegian      &1&Rana Blad& 0.51        & 0.36 &  0.64 & 0.75 & 0.80& 13\% \\
Icelandic     &2 &Fréttablaðið & 0.83        & 0.41 & 0.63 & 0.72 & 0.76& 6\% \\

Estonian        &3&Eesti Päevaleht& 0.67        & 0.59 & 0.78 & 0.85 & 0.90 & 3\%\\
Greek           &3&Kathimerini& 0.61        & 0.56 & 0.85 & 0.94 & 0.97 & 16\%\\
Hebrew    &3&Hadashot& 0.84  &  0.49& 0.74& 0.84 &  0.89 & 26\%\\

Italian         &4&Stampa Sera& 0.79        & 0.36 & 0.60 &  0.69 & 0.74 &  3\%\\

Polish          &4&Dziennik Polski& 0.72        & 0.38 & 0.58 & 0.65 & 0.70 & 30\%\\
%\midrule
%English          &5&CNN/Daily Mail& 0.90        &0.13  & 0.52 & 0.72 & 0.81 &0\%\\
\bottomrule
\end{tabular}
\caption{\textbf{Characteristics of the summaries collected across languages.} $^*$Res. indicates the language resource level, as categorized by \cite{joshi2020state}, showing that we cover languages with varying levels of resources. $^{**}$Comp. stands for Compression rate
 reflecting the document-to-summary length ratio~\citep{grusky2018newsroom}. $^{***}$\%Multi-doc represents the portion of the summaries which span multiple documents.
 Front-page teasers represent high level abstractivity across metrics. %In parentheses, a comparison to the only currently available resource in each language (MassiveSum for X and Multiling for y)
}
  \label{tab:data_info}
  \end{table*}

We show that front-page teasers are a wide spread phenomenon by collecting summary-article pairs in seven languages from diverse families as shown in Table~\ref{tab:data_info}.

We leverage digitization efforts to select a newspaper title for each language, ensuring we can access full scans and manually navigate between pages. We then collect summary-article pairs by manually matching front-page teasers to their corresponding articles, until we reach 30 summaries per language, following \citet{shaib2024much} who found that such sample size could be sufficient to compare model performance on news summarization.  

Matching teasers to articles enables systematic data collection for a wide range of languages. On average, nine issues per title are needed to collect 30 samples, highlighting both the abundance of untapped, naturally annotated data and the manageable effort required to gather sufficient data. Moreover, all newspaper titles featured multi-document summarization, when the teaser summarized several articles. Appendix~\ref{sec:appendixn} details   titles and dates.

% \noamd{missing like this concluded in a small scale multilingual data. in the next section we...}

In Table~\ref{tab:data_info}, we report the level of abstraction in the collected data using the common novel n-gram ratio \cite{narayan2018don}, which measures the percentage of n-grams that appear in the \fpt{} but not in the corresponding article. We also report the compression ratio, comparing the length of the teaser to that of the article. In the next section we use this data to evaluate the performance of state-of-the-art LLMs in summarizing diverse languages against organic, challenging data.

%In Table~\ref{tab:data_info}, we compare \fpt{} data to the widely used CNN/Daily Mail dataset \cite{cheng2016neural, hermann2015teaching}. We find that our data shows higher novel unigram and bigram ratios across all languages, and surpasses CNN/Daily Mail in trigram and four-gram novelty in most cases. In the following section, we use this data to evaluate the performance of state-of-the-art LLMs in summarizing diverse languages against organic, challenging data.

%\subsection{\FPT{} Enables Model Comparison}
%\gabis{I much prefer a dedicated and more conventional Results section instead of this subsection, whose title is too vague IMO.}\noamd{we can try}

\section{Experiments}
\label{sec:Experiments}
%\gabis{Maybe this section can actually go before the Hebrew dataset} \noamd{now thinking about it - maybe אthis should be a part of section 4}

We use the manually collected \fpt{} data to evaluate the performance of state-of-the-art LLMs on news articles in seven languages. %We aim to asses if this data presents a challenge to the models. 
Below, we present the metrics we use to assess summary quality, then describe our use of LLM-as-a-judge to evaluate some of these aspects. We then report our findings, presented in Table~\ref{tab:models}.

\subsection{Metrics and LLM-as-a-Judge}
%We aim to evaluate both the LLMs' capabilities in abstractive text generation across different languages and the extent to which the generated summaries align with the gold references, the \fpt{}.

We report ROUGE-1, ROUGE-2, and ROUGE-L \cite{lin2004rouge} to measure overlap between generated summaries and gold references, a practice that remains common in summarization evaluation \cite{gao-etal-2025-mixed,li2024coverage}. ROUGE is limited to textual overlap so we additionally report BERTScore \cite{zhang2019bertscore} to capture semantic similarity.

To complement 
traditional metrics, which may miss key aspects of information overlap \cite{deutsch2021understanding}, we incorporate three additional evaluation metrics to complement them: \emph{coherence} and \emph{consistency}, which were also used in the human evaluation process, and \emph{coverage}, which assesses the extent to which the predicted summary captures the information presented in the reference summary \cite{liu2022revisiting}. These qualities can help distinguish between a model's ability to generate abstractive summaries in a given language (as measured by coherence and consistency) and its ability to identify the key information.

% To complement 
% traditional metrics, which may miss key aspects of information overlap \cite{deutsch2021understanding}, we incorporate three additional evaluation metrics to complement them: (1) \emph{coherence}, which measures how well the sentences in the predicted summary fit together and whether they sound natural \cite{wang2023chatgpt}; (2) \emph{consistency} - which measures whether the facts in the predicted summary are consistent with the facts in the original article \cite{wang2023chatgpt}; and (3) \emph{coverage} - which assesses the extent to which the predicted summary covers the information presented in the reference summary \cite{liu2022revisiting}. These qualities can help distinguish between a model's ability to generate abstractive summaries in a given language (as measured by coherence and consistency) and its ability to identify the key information, as highlighted by the \fpt{}'s professional writer (measured by coverage).

Ideally, we would have employed human evaluators to evaluate these properties, but this is infeasible due to limited availability of annotators in low-resource languages \cite{bhat2023large}, therefore we rely on LLM-as-a-judge, an evaluation approach which has garnered considerable traction in recent years ~\cite{zheng2023judging}. ~\citet{hada2024metal} has recently established that in the context of summarization, evaluation using a strong model and a detailed prompt aligns with human annotations. However, we note some works warn that LLM evaluation might be skewed in favor of model generated text \cite{forde2024re}, thus showing optimistic results, especially in low resource languages. To make the evaluation more reliable, we use a reference-based method. Additionally, we use GPT-4o solely as the evaluation judge and not as one of the summarization models.

\subsection{Results}

\begin{table*}[tb!]
\centering
\centering
\resizebox{\textwidth}{!}{%
\begin{tabular}{l l c c c c c c c}
\toprule
\textbf{Language} & \textbf{Model} & \multicolumn{3}{c}{\textbf{ROUGE}} & \textbf{BERTscore} & \multicolumn{3}{c}{\textbf{LLM-as-a-Judge (1-5 scale)} $\uparrow$} \\
\cmidrule(lr){3-5} \cmidrule(lr){7-9}
& & 1 & 2 & L & & Coherence& Consistency & Coverage \\

\midrule
Norwegian 
       & Mixtral&0.30&0.08&0.27&0.71&4.00&4.33&2.70\\
        & Llama&\textbf{0.33}&\textbf{0.11}&\textbf{0.30}&\textbf{0.72} &4.41&4.7&2.70\\
         &  DeepSeek-R1&0.26&0.06&0.24&0.70&\textbf{4.52}&\textbf{4.78}&\textbf{2.74}\\
\midrule
Icelandic 
       & Mixtral&0.19&0.04&0.17&0.68&3.08&3.38&2.23\\
       & Llama &\textbf{0.27}&\textbf{0.11}&\textbf{0.25}&\textbf{0.72}&4.46&4.62&\textbf{2.92}\\
      &  DeepSeek-R1&0.21&0.06&0.18&0.69&\textbf{4.58}&\textbf{4.69}&2.62\\
\midrule
Estonian 
       & Mixtral&0.09&0.02&0.08&0.63&2.65&2.65&2.7\\
        & Llama &\textbf{0.11}&\textbf{0.05}&\textbf{0.11}&0.64& 4.70&4.7&3.09\\
        &  DeepSeek-R1&0.07&0.02&0.07&0.64&\textbf{4.78}&\textbf{4.91}&\textbf{3.35}\\
\midrule
Greek 
       & Mixtral&0.14&0.03&0.12&0.67&3.93&4.19&2.63\\
        & Llama&\textbf{0.16}&0.04&\textbf{0.14}&\textbf{0.68}&\textbf{4.48}& 4.48& 2.74\\
        &  DeepSeek-R1&0.15&0.04&0.13&0.67&4.44&\textbf{4.74}&\textbf{3.00}\\
\midrule
Hebrew 
       & Mixtral&0.14&0.04&0.13&0.68&2.86&2.93 &1.93\\
       & Llama &\textbf{0.21}&\textbf{0.09}&\textbf{0.20}&\textbf{0.70}&3.96&4.18&\textbf{2.50}\\
       &  DeepSeek-R1&0.13&0.03&     0.12&  0.68& \textbf{4.54}& \textbf{4.82}&2.25\\
\midrule
Italian 
       & Mixtral&0.28&0.09&0.25&0.69&4.40& 4.36& \textbf{2.84}\\
        & Llama&\textbf{0.30}&\textbf{0.10}&\textbf{0.28}&0.69&4.48&4.64& 2.76\\
        &  DeepSeek-R1&0.24&0.05&0.21&0.68&\textbf{4.64}&\textbf{4.88}&2.80\\
       \midrule
Polish 
       & Mixtral&0.21&0.07&0.20&0.68&4.12&4.08&2.32\\
       & Llama&\textbf{0.23}&\textbf{0.10}&\textbf{0.21}&0.68&3.96&3.84&2.12\\
       &  DeepSeek-R1&0.17&0.05&0.16&0.67&\textbf{4.36}&\textbf{4.36}&\textbf{2.36}\\
\bottomrule
\end{tabular}
}
\caption{\textbf{Model performance on our multilingual data collection in a three shot setting.} The best number per metric for each language is highlighted in bold. While some models produce \emph{coherent} text which is \emph{consistent} with the content of the article, all models struggle with fully \emph{covering} the information in the gold summary.}   
\label{tab:models}
\end{table*}
We evaluate three models: DeepSeek-R1 \cite{guo2025deepseek}, Llama 3.3 70B \cite{llama3modelcard} and Mixtral-8x7b \cite{jiang2024mixtral} and report results on three shot-settings. Temperature is set to 0 to ensure consistent and deterministic outputs. For LLM-as-a-judge we use GPT-4o \cite{hurst2024gpt}. We present results in Table~\ref{tab:models}. Below we discuss several observations.

\paragraph{Generated summaries miss aspects of the gold summary.} 
We observe differences between consistency and coverage across all models and languages. This suggests that while the generated summaries are grounded in the source text (i.e., consistent), they often focus on different content than what appears in the \fpt{} (i.e., lower coverage). There is no substantial variation in performance in this aspect across models within the same language: the average difference in coverage between the best and worst performing models is only 0.345 out of 5. A similar pattern is observed in BERTScore, where the largest difference across models in the same language is just 0.04, suggesting that all models exhibit similar limitations in fully capturing the content of the gold summaries. One possible explanation is that \fpt{} differ from previously used news summaries, which are typically based on web articles and may present a new challenge for LLMs. %Performance in this aspect can be improved by providing few-shot in-context examples from our data, with few-shot settings consistently yielding higher BERTscores compared to zero-shot settings in our experiments \gabis{where do we see this? where did we report zero-shot performance?}.

\paragraph{There is bigger difference in model performance in lower resource languages.}
Most models seem to generate coherent text which is consistent with the source documents in all languages, with coherence and consistency scores exceeding 4.3 out of 5 and showing similar levels across all languages. However, we observe larger performance gaps between models on lower resourced languages. For example, in Icelandic, the gaps between the best and worst performing models in coherence and consistency are 1.5 and 1.3, respectively; for Estonian, the differences are 2.13 and 2.26; and for Hebrew, 1.68 and 1.89. In contrast, the highest-resource languages in our study, Italian and Polish, show much smaller gaps: 0.24 and 0.52 for Italian, and 0.4 and 0.52 for Polish. These findings highlight the need for high-quality data in lower-resource languages to enable meaningful model comparisons.
Interestingly, Norwegian, the lowest-resource language in our work, scores highly across all metrics and models. Further experiments are needed to determine whether this reflects genuinely better performance or whether the metrics are less reliable for low-resource languages like Norwegian.

Although these results offer a preliminary indication of the \fpt{} potential to assess multilingual summarization capabilities, a more robust model evaluation and further training could require larger-scale datasets, beyond what can feasibly be collected through manual annotation. Thus, in the next section, we present an automated method for collecting such data at scale.

\section{Automatic Extraction of  Teaser Summaries and Corresponding Articles}
\label{sec:Implementation}

We describe an automatic process for building a large-scale summarization dataset. %Applied to a single Hebrew newspaper title, this process yields over 7,000 summary-articles pairs, nearly half involving multiple articles per teaser. 
We begin by detailing the automated two-step process: identifying front-page teasers and matching them to their corresponding articles. This is followed by an evaluation of the method, assessing its practicality and suitability for low-resource settings. Finally, we introduce our dataset, \name{}, created by applying this approach to a single Hebrew newspaper title, yielding over 7,000 summaries, nearly half involving multiple articles per teaser.

%We begin by describing our dataset, \name{}. Then detailing the automated two-step process: identifying front-page teasers and matching them to their corresponding articles. Finally, we evaluate the quality of the resulting dataset.
% \usepackage{booktabs}

\subsection{Automatically Identifying Teasers and Matching them to Articles}

We describe how we automate the collection of \fpt{} and the process of matching them to the full articles they reference. We examine several methods suited to varying degrees of resource availability.

\paragraph{Identifying Teasers.} Front-page teasers often include key phrases that indicate that the full story appears elsewhere in the issue. Thus, to identify them automatically, we use a rule-based approach that searches for terms specific to each newspaper title. To ensure we include only teasers that function as summaries, we collect additional phrases that signal continuation and use them to filter out non-summary content.

\paragraph{Matching between \fpt{} and corresponding articles.} Automatically determining which article supports a teaser is challenging. While teasers often point to the relevant page (or pages), the specific article or articles on that page must still be identified.

We develop several methods to perform the matching automatically, spanning from those requiring minimal resources, suitable for low-resource settings, to those utilizing LLMs without requiring any human annotations. We test the proposed methods on Hebrew, which poses unique challenges for LLMs as a morphologically rich language, often resulting in ambiguous structures~\cite{tsarfaty2020spmrl}.
%\gabis{can save space by removing enumeration and just putting new lines between each of the methods}

\textbf{TF-IDF (low resource)}: We follow \citet{ghalandari2020largescalemultidocumentsummarizationdataset} that used Term Frequency-Inverse Document Frequency (TF-IDF) to pair news articles with summaries. We generate vector representations for all articles and teasers by computing TF-IDF across the entire corpus, capturing both term frequency and rarity. We then calculate cosine similarity between each teaser and all potential article matches -- i.e., the articles on the corresponding page. We define a similarity threshold, based on manually annotated data (50 samples), above which articles are linked to the teaser. 

\textbf{Finetuned Sentence Transformer (medium resource)}: We use a SentenceTransformer model \cite{reimers2019sentencebertsentenceembeddingsusing}  to compute similarity scores between teasers and articles by measuring the cosine similarity between their embeddings. A threshold is then applied to determine matches. %To improve performance, we first fine-tuned the model on our annotated pairs, following the SetFit approach \cite{tunstall2022efficient}, which requires only a small number of examples.

\textbf{Zero shot classification (medium to high resource)}: We prompt an LLM to determine for each teaser-article pair whether the teaser summarizes the text. The instruction was given in English, and the text in the original language.

We apply all these methods to a Hebrew newspaper collection. In the next section, we report their quality assessment.

% % Please add the following required packages to your document preamble:
% % \usepackage{booktabs}
% \begin{table}[]
% \small

% \begin{tabular}{@{}lllr@{}}
% \toprule
% \textbf{Dataset}    &\textbf{\#Samples} & \textbf{Shape} & \textbf{\# Multi-doc}   \\ \midrule
% MassiveSumm & 102,961   &   Paragraph &\textcolor{red}{X}          \\
% HeSum      & 10,000    &   Paragraph & \textcolor{red}{X}         \\
% \name{}       & \numsamples{}  & Diverse      & \textbf{\nummulti{}}         \\ \bottomrule
% \end{tabular}

%  \caption{Comparing our dataset to the available Hebrew summarization datasets: Hesum \cite{mondshine2024hesum} and MassiveSumm \cite{varab-schluter-2021-massivesumm}}
%    \label{tab:multidoc}

% \end{table}

\begin{table}[tb!]
\centering
\centering
\resizebox{\columnwidth}{!}{%
\begin{tabular}{llc}
\toprule
\textbf{Error Type} &  \textbf{Percentage}&\textbf{Category}  \\
\midrule
Segmentation Error & 66\% &False negative \\
Length           & 20\% &False positive \\
OCR Noise             & 13\%   &False negative \\
\bottomrule
\end{tabular}
}
\caption{Error analysis of teaser identification using a rule-based approach. }
\label{tab:error_analysis}

\end{table}

% \usepackage{booktabs}
\begin{table}[tb!]
\centering
\small
\resizebox{\columnwidth}{!}{%
\begin{tabular}{@{}lllll@{}}
\toprule
\textbf{Method}      & \textbf{Acc.} & \textbf{Prec.} & \textbf{Rec.} & \textbf{F1} \\ \midrule
TF-IDF                & 86            & 93             & 75           & 83          \\
Sentence Transformer & 81            & 82             & 77           & 80          \\
Zero-shot (Llama-3.3 70B)           & \textbf{90}            & \textbf{95}             & \textbf{83}           & \textbf{88}          \\ \bottomrule
\end{tabular}%
}
\caption{Evaluation of automatic teaser-to-article matching. We report accuracy (Acc), precision (Prec), recall (Rec) and F1 score on the Hebrew dataset.}
\label{tab:match}
\end{table}

\subsection{Dataset Quality and Error Analyses}
We manually evaluate the automatic process described above finding that it provides a practical option for low-resource settings. All annotations are carried out by native speakers of Hebrew.

\paragraph{Identifying teasers using a rule-based approach yields strong performance.} 
After collecting \fpt{} using key phrases, we aim to assess how well this method identifies texts that function as actual summaries. We manually tag 102 texts appearing on front pages to determine whether they could be considered summaries. We then compare these annotations to the results of our rule-based method, which classifies a text as a teaser if it contains a key phrase from a predefined, language-specific, list.
We analyze the types of errors in Table~\ref{tab:error_analysis}.

We achieve precision of 95\% and recall 85\%. The error analysis reveals that 80\% of the errors were false negatives (i.e., teasers that were not identified as summaries), which affects the dataset’s scale and coverage but not its quality. These errors are primarily caused by segmentation errors, specifically cases where the \fpt{} text is unavailable to the automatic process because it is marked as an image caption but human annotators can still identify it. Another source of error is OCR noise, where key phrases appear in the \fpt{} but are too distorted to be correctly detected. 20\% of the errors are false positives (i.e., texts which were incorrectly marked as teasers), in all of which headers are incorrectly marked as teasers because they follow the teaser format, using the same key word to point to a different page.

% \begin{table}[tb!]
% \centering
% \small
% \begin{tabular}{llc}
% \toprule
% \textbf{Error Type} &  \textbf{Percentage}&\textbf{Category}  \\
% \midrule
% Segmentation Error & 10 &False negative \\
% Length           & 3 &False positive \\
% OCR Noise             & 2   &False negative \\
% \bottomrule
% \end{tabular}
% \caption{Error analysis of teaser identification using a rule-based approach.\noamd{}}
% \label{tab:error_analysis}
% \end{table}

%Out of the 102 texts we annotated, the automatic process incorrectly tagged five as summaries. Four of these were very short, spanning just a few words, and were therefore not considered summaries in the manual annotation. \gabis{This paragraph is heavy with numbers, consider moving them all to a table, perhaps with some concrete examples for the errors? Also, I don't like line breaks within paragraph environments.}

% \begin{table}[tb!]
% \centering
% \small
% \begin{tabular}{llc}
% \toprule
% \textbf{Error Type} &  \textbf{Count}  \\
% \midrule
% Related articles             & 16    \\
% OCR noise & 13 \\
% Others           &  10 \\

% Length           & 2 \\

% \bottomrule
% \end{tabular}
% \caption{Error analysis of teaser identification using a rule-based approach.\gabis{here too, percent would be more meaningful IMO, also maybe add an example? or is too confusing since it's not in English}}
% \label{tab:error_analysisTFIDF}
% \end{table}

\paragraph{In matching front-page teasers to articles, the zero-shot approach performs best, while TF-IDF presents a viable alternative.} To evaluate the performance of the three proposed methods for matching teasers to summaries -- TF-IDF, Sentence Transformers and zero-shot -- we manually annotate 50 teasers along with all articles on the pages to which they point. This results in 325 teaser-article pairs, with an average of 6.5 article candidates per teaser. Each pair is annotated to determine whether the teaser can be considered a summary of the article. The annotators overlapped on 110 pairs, yielding a Cohen’s kappa~\cite{cohen1960coefficient} of 0.88, indicating very strong agreement. We then compare our annotations to the classifications produced by each method, with the results presented in Table~\ref{tab:match}.

The best performance is observed in the zero-shot setting, using instruct tuned Llama-3.3 70B~\cite{llama3modelcard}. TF-IDF also performs well, outperforming the sentence transformer on most metrics, while requiring minimal computation. %However, it also resulted in a higher false negative rate, failing to identify some related articles. \gabis{Where do we see this?}

To better understand the capabilities of the low-resource method, we provide an error analysis of the TF-IDF approach in Table~\ref{tab:error_analysisTFIDF}. The most common error occurs when the article is related to the teaser but not directly supported by it, for example, when both refer to the same event, but the teaser does not explicitly mention it. Other errors occur on noisy samples affected by OCR issues, which a language model is able to handle more robustly, and when the teaser text is short.

\subsection{\name{}: A Real-World Hebrew Summarization Corpus}
Our Hebrew dataset, collected using the above method, features a rich variety of summarization shapes and is naturally multi-document. 

We apply our method to the ``Hadashot'' newspaper, using issues published between 1984 and 1993, available via the National Library of Israel. From a single title, we extract \numsamples{} samples, of which \nummulti{} are multi-document (i.e., the summary refers to more than one article).
To the best of our knowledge, this is the first available dataset in Hebrew to support multi-document summarization. 
%Moreover, it includes a variety of summary shapes ranging from single sentences to bullet points to one or more paragraphs. 
In table \ref{tab:multidoc}, we compare our data to the available datasets in Hebrew. A detailed summarization datacard and data statistics are provided in Tables~\ref{tab:datacard} and~\ref{tab:summary_stats}.
As the human annotation found length serves as an indicator of quality, we also report the number of teasers per length category.

%\noamd{As the human annotation found length serves as an indicator of quality, we also report the number of teasers per length category. Most of the teasers in our dataset, \textasciitilde6k samples are over 25 words.}

The contribution of this dataset is two-fold. First, it serves as an example that collecting summarization corpora is feasible, with relatively few resources. Second, the resulting dataset can be used to advance the state of the art in Hebrew NLP. Large-scale data collection remains important for Hebrew, as it enables robust evaluation and supports fine-tuning, for example, to improve morphological understanding \cite{mondshine2024hesum}. 

% We will make the dataset publicly available using our code, and we provide additional metadata, including summary shape and whether the example is single or multi-document. 
We will make the dataset publicly available using our code. To account for OCR errors, we provide a corrected version of the data, achieved by performing prompt-base OCR post correction~\cite{thomas2024leveraging}. 
\begin{table}[tb!]
\centering
\resizebox{\columnwidth}{!}{%
\begin{tabular}{@{}lllr@{}}
\toprule
\textbf{Dataset}    &\textbf{\#Samples} & \textbf{Shape} & \textbf{Multi-doc}   \\ \midrule
MassiveSumm & 102,961   &   Paragraph &\textcolor{red}{x}          \\
HeSum      & 10,000    &   Paragraph & \textcolor{red}{x}         \\
Ours       & \numsamples{}  & Diverse      & \textbf{\nummulti{}}         \\ \bottomrule
\end{tabular}
}
 \caption{Comparing our dataset to the available Hebrew summarization datasets: Hesum \cite{mondshine2024hesum} and MassiveSumm \cite{varab2021massivesumm}.}
   \label{tab:multidoc}

\end{table}                                              
\section{Related work}
\label{sec:related}

Many summarization datasets rely on the news domain~\cite{dahan-stanovsky-2025-state} but they are based on web content. Common methods for constructing such datasets include using the article title as a summary~\cite{narayan2018don}, extracting the first bolded sentence~\cite{cheng2016neural, hermann2015teaching}, or leveraging social media metadata~\cite{grusky2018newsroom}. A few works have focused on  identifying summaries and matching them to full articles, which requires an automatic matching process. For example, \citet{ghalandari2020largescalemultidocumentsummarizationdataset} linked articles from Wikipedia’s Current Events Portal to news articles covering the events.

In contrast to web-based news, newspapers have received limited attention in NLP. However, several works have explored named entity recognition on historical newspapers ~\cite{boros-etal-2020-alleviating,ehrmann2020introducing}, while others have leveraged NLP methods to analyze historical texts \cite{borenstein2023measuring,candela2022discovering}.

% \section{The Hebrew Dataset}
% \label{sec:HebrewData}
% \input{latex/sections/HebrewDataset}                                    
\section{Conclusion}
\label{sec:conclusion}
We presented a simple method for collecting high-quality summarization data from printed newspapers, demonstrating its suitability for languages with varying levels of resources. The resulting data is abstractive, supports multi-document summarization, and enables evaluation of model capabilities. Using this method, we created a new summarization dataset for Hebrew, \name{}, which we release  using our code.

\section*{Limitations}
\label{sec:Limitations}
Although this work supports summarization data collection across diverse languages, we wish to acknowledge several limitations. First, while the method benefits languages with limited online presence, it is less applicable to languages that fell out of use before the 20th century, as front-page teasers only became widespread during that period. Additionally, the quality of the collected data is constrained by OCR errors, which may affect downstream results. For example, highly noisy text can hinder the effectiveness of methods like TF-IDF and may distort evaluation metrics such as novel n-gram overlap.                                    

\section*{Ethical Considerations}
\label{sec:Ethics}
Digitized newspapers are provided by national libraries for research purposes, but their content may still be subject to copyright restrictions. To respect these restrictions, we do not publish the collected data online. However, the Hebrew dataset can be reconstructed using the library’s API and terms of service, and the multilingual data can be recreated using the list of titles and publication dates provided in the appendix.

\bibliography{latex/custom}
\appendix

\section{Newspaper Use}
\label{sec:appendixn}

We provide links to the original newspapers from which the \fpt{} examples in Figure~\ref{fig:example_multiling} were taken and their translations, along with the issue dates used to collect \fpt{} for Section~\ref{sec:method_new}. We also provide a list of key phrases used to identify teasers in various languages. 
\subsection{Picture Origin:}
\begin{enumerate}
    \item \href{https://www.nb.no/items/cbb9204728ff5517bb6e3b0761414690?page=15}{Rana Blad (Norway)}
    \item \href{http://www.archiviolastampa.it/component/option,com_lastampa/task,search/mod,libera/Itemid,3,/action,viewer/page,0001/articleid,1296_02_1991_0226_0001/}{Stampa Sera (Italy)}
    \item \href{https://timarit.is/page/2125475#page/n0/mode/2up}{Fréttablaðið (Iceland)}
\end{enumerate}

\subsection{Translations:}
Translations of the teasers shown in Figure~\ref{fig:example_multiling} are provided to illustrate the data quality, generated using GPT-4o.
\begin{enumerate}
\item Today the National Ski Championship opens, and the first day may belong to the recruits. Elin Nilsen from B\&Y IL, and Inger Lise Hegge from Henning, are both recruits who earlier in the season have challenged the national team skiers. The women will race 10 kilometers today, while the men will do 30 kilometers. In addition, there is the combined ski jumping and team jumping events.
\item Hopes of forming a national government in Macedonia diminished significantly yesterday, as an Albanian political party demands that attacks on armed Albanian residents in Kosovo be halted before talks begin.
\item Alessandra Casella, 27 years old, is the co-star of Fredi, currently showing at the Alfieri Theater in Turin. After her experience at the Actor’s Studio and lots of theater, she would like to work more in cinema, but as she told us (interview on page 27), ``in Italy, times are tough — directors prefer fashion models.''
\end{enumerate}
\subsection{Dates for collection:} For the data collection in Section~\ref{sec:method_new}, we randomly select a starting date and review all subsequent issues of the title, following the library’s order, which was sometimes not chronological. We present dates here.

\begin{enumerate}
    \item \textbf{Rana Blad (Norway)}: 01.02.1990-10.02.1990
    \item \textbf{Kathimerini (Greece)}: 01.12.2010, 02.11.2010, 01.10.2010, 01.09.2010, 17.08.2010, 01.07.2010, 01.05.2010, 01.06.2010
    \item \textbf{Stampa Sera (Italy)}: 09.10.1991-18.10.1991
    \item \textbf{Eesti Päevaleht (Estonia)}: 01.01.2018, 31.01.2018, 02.02.2018, 03.03.2018, 02.03.2018, 01.04.2018, 02.04.2018, 01.05.2018
    \item \textbf{Fréttablaðið (Iceland)}: 23.04.2001, 24.04.2001, 26.04.2001, 30.04.2001, 02.05.2001, 03.05.2001, 04.05.2001, 07.05.2001, 08.05.2001, 09.05.2001, 10.05.2001, 11.05.2001
    \item \textbf{Hadashot (Israel)} 01.01.1992 - 08.01.1992
    \item \textbf{Dziennik Polski (Poland)}: 01.01.2002, 03.01.2002, 05.01.2002, 07.01.2002, 08.01.2002, 09.01.2002, 10.01.2002
\end{enumerate}

\subsection{Key phrases}
For the manual collection we used the following key phrases to identify teasers:
\begin{itemize}
    \item \textbf{Norwegian}:Side. (``Page'')
    \item \textbf{Icelandic}:Bls. (abbreviation for ``Page'')
    \item \textbf{Estonian}:LK. (abbreviation for ``Page'')
    \item \textbf{Greek}: Sel, written in Greek script. (abbreviation for ``Page'')
    \item \textbf{Hebrew}: Am, written in Hebrew script. (abbreviation for ``Page'') 
    \item \textbf{Italian}: Pag. (abbreviation for ``Page'')
    \item \textbf{Polish}: Szczegóły and str. (Details and abbreviation for ``Page'', respectively)

\end{itemize}

\section{Teasers Examples}
\label{sec:appendixe}

We present here translated examples of teasers from our dataset. 
\paragraph{Under 25 words:} ``Children Destroyed 2.6 Million Shekels in Search of Smurf Stickers.''
\paragraph{between 25 and 50 words:} ``Thousands Spent the Holiday Away from Home: 25,000 Travelers Flocked to the Beaches of Eilat. The hotels, hostels, and especially the beaches were filled to capacity. The beaches of the Sea of Galilee were also crowded, particularly Tzemach Beach, where a rock festival was taking place. About ten thousand people passed through the Taba terminal and traveled to Sinai.''
\paragraph{between 50 and 100 words:} ``Driver Who Ran Over Child During 'Road Roulette' Acquitted. Boris Eligolashvili of Ramla, who was accused of causing the death of 11-year-old Baruch Oren from Haifa, was acquitted yesterday in the Haifa Traffic Magistrate’s Court. The judge ruled that the accident was not caused by his negligence. Baruch Oren was killed on March 17, 1989, on the Haifa–Tel Aviv highway, near Kiryat Shprintzak in Haifa. Oren’s friends said that he was waiting in the median strip of the road, and when Eligolashvili’s car was about a hundred meters away, he jumped onto the road and stood in front of it with his arms folded.''
\paragraph{over 100 words:} ``120,000 Vaccinated Against Polio in Hadera Governorate as of This Morning; Two Additional Cases Suspected. The Ministry of Health is launching an unprecedented campaign this morning, during which 120,000 people in Hadera Governorate will be vaccinated against polio. These are individuals aged 0 to 34 years. The vaccine, which protects recipients from the virus, will be administered orally using the Sabin vaccine. Pregnant women and infants who have never been vaccinated will receive an injection of the Salk vaccine. Today, about a thousand infants and all pregnant women in Hadera Governorate will be vaccinated. The day after tomorrow and on Friday, the Sabin vaccine will be administered to about 50,000 kindergarten-age children and students in the education system. Starting next Sunday, the adult population and children below kindergarten age will be vaccinated. The Ministry of Health announced yesterday two new suspected cases of polio: a 27-year-old resident of Hadera and a young man, about 24 years old, from Zichron Yaakov. The condition of the nine-month-old baby from Kiryat Gat, who was brought to Barzilai Hospital three weeks ago with suspected polio, has improved.''

\section{Human annotation}
\label{sec:appendixh}
We follow~\citet{fabbri2021summeval} annotation guidelines.

``In this task you will evaluate the quality of summaries written for a news article.
To correctly solve this task, follow these steps:
\begin{enumerate}
    \item Carefully read the news article(s), be aware of the information it contains.
    \item Read the proposed summary
    \item Rate each summary on a scale from 1 (worst) to 5 (best) by its relevance, consistency, fluency, and coherence.

\end{enumerate}

Definitions:

Relevance:
The rating measures how well the summary captures the key points of the article.
Consider whether all and only the important aspects are contained in the summary. When multiple articles are presented, the score should account for the most important information across all sources.

Consistency:
The rating measures whether the facts in the summary are consistent with the facts in the original article.
Consider whether the summary does reproduce all facts accurately and does not make up untrue information.

Fluency:
This rating measures the quality of individual sentences, are they well-written and grammatically correct.
Consider the quality of individual sentences.

Coherence:
The rating measures the quality of all sentences collectively, to the fit together and sound naturally.
Consider the quality of the summary as a whole.
''

\section{Prompts}
\label{sec:appendixp}
We provide the prompts used in our work. 
\paragraph{Prompt use for summarization task:} We follow \citet{wang2023zero}.     "Please summarize the following text in [LANGUAGE]:

Text: [TEXT]

\paragraph{Prompts use for LLM-as-a-judge:}
\begin{itemize}
    \item \textbf{Coherence:} We follow \citet{wang2023chatgpt}. "Score the following news summarization given the corresponding news with respect to coherence with one to five stars, where one star means “incoherence” and five stars means “perfect coherence”. Note that coherence measures the quality of all sentences collectively, to the fit together and sound naturally. Consider the quality of the summary as a whole."
    \item \textbf{Consistency:} We follow \citet{wang2023chatgpt}. "Score the following news summarization given the corresponding news with respect to consistency with one to five stars, where one star means “inconsistency” and five stars means “perfect consistency”. Note that consistency measures whether the facts in the summary are consistent with the facts in the original article. Consider whether the summary does reproduce all facts accurately and does not make up untrue information."
    \item \textbf{Coverage:} We follow \citet{liu2022revisiting}. "You will receive a reference summary and a candidate summary. Your task is to compare these two summaries and assess the extent to which the candidate summary covers the information presented in the reference summary.
    
    Please indicate your agreement with the following statement: “All of the information in the reference summary can be found in the candidate summary.”
    
    Use the following 5-point scale when determining your response:
    
    1. Strongly Disagree
    
    2. Disagree

    3. Neither Agree nor Disagree
    
    4. Agree
    
    5. Strongly Agree
    
    Reference Summary:{referece}
    
    Candidate Summary: {generated}
    
    Evaluation Form (scores ONLY): - Agreement (1-5):"

\end{itemize}

\paragraph{Prompt use for matching teasers to articles:} "Given the following text and summary, answer with 'Yes' if the text relates to the summary, and 'No' if it does not. Do not provide explanations. Only output 'Yes' or 'No'."

\section{Data Statistics}
\label{sec:appendixs}
\begin{table}[tb!]
\resizebox{\columnwidth}{!}{%
\begin{tabular}{|p{7.5cm}|}

\hline
\textbf{Summarization Data Card}                                                                                      \\ \hline
\textbf{\underline{Sample information:}}      
\\
\textbf{Languages:} 
\newline
\textit{Hebrew}                                                                               \\
\textbf{Summary Shape:}
\newline
\textit{Diverse:}
Paragraph: \numparagraph; One-Sentence:\numsentence ; Highlights:\numhighlights 
\\
\textbf{Summary Distribution by Length:}
\newline
0-25: 2031 \newline
25-50: 1160\newline
50-100: 2201\newline
>100: 2382
\\
\textbf{Domain:} 
\newline
\textit{News}                                                                       \\
\textbf{Size:}
\newline
\textit{\numsamples{}}                                                                                \\ \hline
\textbf{\underline{Annotation information:}}                                                                                      \\
\begin{tabular}[|p{7.5cm}|]{@{}l@{}}\textbf{Annotation efforts:} \\ \textit{Automatic}\end{tabular}   \\
\begin{tabular}[|p{7.5cm}|]{@{}l@{}}\textbf{Source of supervision:}\\ \textit{Natural} (summaries created organically)\end{tabular} \\
\begin{tabular}[|p{7.5cm}|]{@{}l@{}}\textbf{Brief description of the summaries' source:} \\ \textit{Newspapers front page teasers}\end{tabular} \\ \hline
\textbf{\underline{Data quality assessment:}}                                                                                     \\
\begin{tabular}[|p{7.5cm}|]{@{}l@{}}\textbf{Abstraction level:} \\ \textit{1-gram ratio}: 0.58 \\ \textit{2-gram ratio:} 0.82\\
\textit{3-gram ratio:} 0.89 \\
\textit{4-gram ratio:} 0.92\end{tabular} \\
\textbf{Compression rate:} 0.84 \\
\textbf{Human evaluation:} \textit{by domain experts}                                                                                              \\ \hline
\textbf{\underline{Availability details:}}                                                                                        \\
\begin{tabular}[|p{7.5cm}|]{@{}l@{}}\textbf{How is the data made accessible:} \\ \textit{URL-based Reconstruction}\end{tabular}    \\
\begin{tabular}[c]{@{}l@{}}\textbf{Copyrights information:} \\ \textit{License}\end{tabular}                                            \\ \hline
\end{tabular}%
}
\caption{Summarization data card.}
\label{tab:datacard}

\end{table}

\begin{table}[h]
\resizebox{\columnwidth}{!}{%
\centering
\begin{tabular}{lcc}
\toprule
\textbf{Statistic} & \textbf{Single-Doc} & \textbf{Multi-Doc} \\
\midrule
\# Teasers & 3,905 & 3,869 \\
Avg. text length (words) & 405 & 312 \\
Avg. summary length (words) & 56 & 102 \\
\bottomrule
\end{tabular}%
}
\caption{Dataset statistics for single and multi-document data. The average number of articles in cluster is 3.4 }
\label{tab:summary_stats}

\end{table}

\begin{table}[tb!]
\centering
%small
\begin{tabular}{llr}
\toprule
\textbf{Error Type} &  \textbf{Percentage}  \\
\midrule
Related articles             & 39\%    \\
OCR noise & 32\% \\
Others           &  24\% \\

Length           & 4\% \\

\bottomrule
\end{tabular}
\caption{Error analysis of the TF-IDF approach to match teasers to articles. Out of 325 pairs we find 41 errors.}
\label{tab:error_analysisTFIDF}
\end{table}

% - name and date of all publications
% - all key words in hebrew
% -prompts

% \section{Prompts}
% \input{latex/sections/Appendix_prompts}
% \label{sec:appendixp}

\end{document}